\documentclass{article}

\usepackage{chieh}                      
\usepackage[final]{neurips_2020}       

\usepackage[utf8]{inputenc} 
\usepackage[T1]{fontenc}    
\usepackage{hyperref}       
\usepackage{url}            
\usepackage{booktabs}       
\usepackage{amsfonts}       
\usepackage{nicefrac}       
\usepackage{microtype}      

\title{Layer-wise Learning of Kernel Dependence Networks}

\author[1]{\thanks{Signifies equal contribution.} Chieh Wu}
\author[1]{$^*$Aria Masoomi}
\author[2]{Arthur Gretton}
\author[1]{Jennifer Dy}

\affil[1]{Department of Electrical and Computer Engineering, Northeastern University}
\affil[2]{Gatsby Computational Neuroscience Unit, University College London}

\begin{document}
\maketitle
\setcitestyle{numbers,square}

\begin{abstract}
We propose a greedy strategy to spectrally train a deep network for multi-class classification. Each layer is defined as a composition of linear weights with the feature map of a Gaussian kernel acting as the activation function. At each layer, the linear weights are learned by maximizing the dependence between the layer output and the labels using the Hilbert Schmidt Independence Criterion (HSIC). By constraining the solution space on the Stiefel Manifold, we demonstrate how our network construct (Kernel Dependence Network or \kc) can be solved spectrally while leveraging the eigenvalues to automatically find the width and the depth of the network. We theoretically guarantee the existence of a solution for the global optimum while providing insight into our network's ability to generalize. This workshop paper is only part one of the full paper. For the full paper, see \url{https://arxiv.org/abs/2006.08539}.

\end{abstract}

\section{Network Model}
Let $X \in \mathbb{R}^{n \times d}$ be a dataset of $n$ samples with $d$ features and let $Y \in \mathbb{R}^{n \times \nclass}$ be the corresponding one-hot encoded labels with $\nclass$ number of classes. Let $\cS$ be a set of $i,j$ sample pairs that belong to the same class. Its complement, $\cS^c$ contains all sample pairs from different classes. Let $\odot$ be the element-wise product. The $i^{th}$ sample and label of the dataset is written as $x_i$ and $y_i$. $H$ is a centering matrix defined as $H = I_n - \frac{1}{n} \mathbf{1}_n \mathbf{1}_n^T$ where $I_n$ is the identity matrix of size $n \times n$ and $\textbf{1}_n$ is a vector of 1s also of length $n$. Given $H$, we let $\Gamma = HYY^TH$.

We denote the network linear weights as $W_1 \in \mathbb{R}^{d \times q}$ and $W_l \in \mathbb{R}^{m \times q}$ for the 1st layer 
and the $l^{th}$ layer; assuming $l > 1$. The input and output at the $l^{th}$ layer are $R_{l-1} \in \mathbb{R}^{n \times m}$ and $R_{l} \in \mathbb{R}^{n \times m}$, i.e., given $\af:\mathbb{R}^{n \times q} \rightarrow \mathbb{R}^{n \times m}$ as the activation function, $R_l = \af(R_{l-1}W_l)$. For each layer, the $i^{th}$ row of its input $R_{l-1}$ is $r_i \in \mathbb{R}^m$ and it represents the $i^{th}$ input sample. We denote $\mathcal{W}_l$ as a function where $\mathcal{W}_l(R_{l-1}) = R_{l-1}W_l$; consequently, each layer is also a function $\fm_l = \af \circ \mathcal{W}_l$. By stacking $L$ layers together, the entire network itself becomes a function $\fm$ where $\fm = \fm_L \circ ... \circ \fm_1$. Given an empirical risk $(\hsic)$ and a loss function $(\mathcal{L})$, our network model assumes an objective of
    \begin{equation}
    \underset{\fm}{\min} \hspace{0.3cm} \hsic(\fm)
    =
    \underset{\fm}{\min} \hspace{0.3cm} \frac{1}{n} \sum_{i=1}^n \mathcal{L}(\fm(x_i), y_i).
     \label{eq:basic_empirical_risk}
    \end{equation}
     Notice that \kc fundamentally models a traditional fully connected multilayer perceptron (MLP) where each layer consists of linear weights $W_l$ and a activation function $\af_l$. We propose to solve \eq{eq:basic_empirical_risk} greedily; this is equivalent to solving a sequence of \textit{single-layered} networks where the previous network output becomes the current layer's input. At each layer, we find the $W_l$ that maximizes the dependency between the layer output and the label via the Hilbert Schmidt Independence Criterion (HSIC) \cite{gretton2005measuring}:
\begin{equation}
    \max_{W_l} \Tr \left(
        \Gamma \:
        \left[
        \af(R_{l-1}W_l) \af^T(R_{l-1}W_l)
        \right]
        \right)  \quad \st W_l^TW_l=I.
        \label{eq:main_obj}
\end{equation}
While Mean Squared Error (MSE) and Cross-Entropy (CE) have traditionally been used for classification, HSIC is instead chosen because it solves an underlying prerequisite of classification, i.e., learning a mapping for $X$ where similar and different classes become distinguishable.    However, since there are many notions of similarity, it is not always clear which is best for a particular situation. \kc overcomes this uncertainty by discovering the optimal similarity measure as a kernel function during training.
To understand how, first realize that the $i,j_{th}$ element of $\Gamma$, denoted as $\Gamma_{i,j}$, is a positive value for samples in $\cS$ and negative for $\cS^c$. By defining a kernel function $\kf$ as a similarity measure between 2 samples, HSIC as \eq{eq:main_obj} becomes a kernel discovery objective written as
\begin{equation}
    \max_{W_l} 
    \sums \Gij \kf_{W_l}(r_i, r_j) 
    -
    \sumsc |\Gij| \kf_{W_l}(r_i, r_j)
    \quad \st W_l^TW_l=I.
    \label{eq:similarity_hsic}
\end{equation}
Notice that the objective uses the sign of $\Gamma_{i,j}$ as labels to guide the choice of $W_l$ such that it increases $\kf_{W_l}(r_i,r_j)$ when $r_i,r_j$ belongs to $\cS$ while decreasing $\kf_{W_l}(r_i,r_j)$ otherwise. Therefore, by finding a $W_l$ matrix that best parameterizes $\kf$, HSIC discovers the optimal pair-wise relationship function $\kf_{W_l}$ that separates samples into similar and dissimilar partitions. Given this strategy, we will formally demonstrate how learning $\kf$ leads to classification in the following sections.

While \kc uses a MLP structure as a basis, we deviate from a traditional MLP in two respects. First,the traditional concept of activation functions is replace by a feature map of a kernel. For \kc, we use the feature map of a Gaussian kernel (\rbfk) to simulate an infinitely wide network. Yet, the kernel trick spares us the direct computation of $\af(R_{l-1}W_l) \af^T(R_{l-1}W_l)$; we instead compute the \rbfk matrix given $\kf(W_l^T r_i, W^T_lr_j) = \text{exp}\{-||W_l^T r_i - W_l^T r_j||^2/2\sigma^2\}$. Second, the constraint $W^TW = I$ is inspired by the recent work on the geometric landscape of the network solutions \cite{poggio2020complexity,li2018measuring,fort2019large}. Their work suggests that the network solution can be represented by a linear subspace where the only the direction of the weights matter and not the magnitude. If the solution indeed lives on a linear subspace independent of their scale, we can exploit this prior knowledge to narrow the search space during optimization specific to the Stiefel Manifold where $W_l^TW_l=I$, rendering \eq{eq:main_obj} solvable spectrally. Consequently, this prior enables us to solve \eq{eq:main_obj} by leveraging the iterative spectral method (ISM) proposed by \citet{wu2018iterative,Wu2019SolvingIK} to simultaneously avoid SGD and identify the network width. Applying ISM to our model, each layer's weight is initialized using the most dominant eigenvectors of
\begin{equation}
    \label{eq:init_phi}
    \mathcal{Q}_{l^0} = R_{l-1}^T (\Gamma - \text{Diag}(\Gamma 1_n))  R_{l-1},
\end{equation}
where the Diag($\cdot$) function places the elements of a vector into the diagonal of a square matrix with zero elements. Once the initial weights $W_{l^0}$ are set, ISM iteratively updates $W_{l^{\mathbf{i}}}$ to $W_{l^{\mathbf{i+1}}}$ by setting $W_{l^{\mathbf{i+1}}}$ to the most dominant eigenvectors of
\begin{equation}
    \mathcal{Q}_{l^i} = R_{l-1}^T (
    \hat{\Gamma}
 - \text{Diag}(\hat{\Gamma} 1_n))  R_{l-1}
    ,
    \label{eq:phi}
\end{equation}
where $\hat{\Gamma}$ is a function of $W_{l^\mathbf{i}}$ computed with $\hat{\Gamma} = \Gamma \odot K_{R_{l-1}W_{l^{\mathbf{i}}}}$. This iterative weight-updating process stops when $\mathcal{Q}_{l^\mathbf{i}} \approx \mathcal{Q}_{l^{\mathbf{i+1}}}$, whereupon $\mathcal{Q}_{l^{\mathbf{i+1}}}$ is set to $\mathcal{Q}_l^*$, and its most dominant eigenvectors $W_{l}^*$ becomes the solution of \eq{eq:main_obj} where $\partial \hsic/\partial W_l = 0$.

ISM solves \eq{eq:main_obj} via the kernel trick directly on an infinitely wide network during training, obtaining $W_l^*$. Once $W_l^*$ is solved, we approximate $\af$ with Random Fourier Features (RFF) \cite{rahimi2008random} to finitely simulate an infinitely wide network and acquire the layer output. This output is then used as input for the next layer. Capitalizing on the spectral properties of ISM, the spectrum of $\mathcal{Q}_l^*$ completely determines the  
the width of the network  
$W_l^* \in \mathbb{R}^{m \times q}$, i.e., $m$ is equal to the size of the RFF, and $q$ is simply the rank of $\mathcal{Q}_l^*$. Furthermore, since \eq{eq:main_obj} after normalization is upper bounded by 1, we can stop adding new layers when the HSIC value of the current layer approaches this theoretical bound, thereby prescribing a natural depth of the network. The resulting network $\fm$ after training will map samples of the same class into its own cluster, allowing the test samples to be classified by matching their network outputs to the nearest cluster center.  
The source code is included in the supplementary materials and made publicly available at \url{https://github.com/endsley/kernel_dependence_network}.

\section{Theoretical Origin of Kernel Dependence Networks}
\textbf{Background and Notations. } 
Let the composition of the first $l$ layers be $\fm_{l^\circ} = \fm_l \circ ... \circ \fm_1$ where $l \leq L$.  This notation enables us to connect the data directly to the layer output where $R_l = \fm_{l^{\circ}}(X)$. Since \kc is greedy, it solves MLPs by replacing $\fm$ in \eq{eq:basic_empirical_risk} incrementally with a sequence of functions $\{\fm_{l^\circ}\}_{l=1}^L$ where each layer relies on the weights of the previous layer. This implies that we are also solving a sequence of empirical risks $\{\hsic_l\}_{l=1}^L$, i.e., different versions of \eq{eq:basic_empirical_risk} given the current $\fm_{l^\circ}$. We refer to $\{\fm_{l^\circ}\}_{l=1}^L$ and $\{\hsic_{l}\}_{l=1}^L$ as the \KS and the \RS.  

Following the concept mentioned above, we guarantee the existence of a solution to reach the global optimum given the theorem below with its proof in App.~A . 

\begin{adjustwidth}{0.2cm}{0.0cm}
\vspace{5px}
\begin{theorem}
\label{thm:hsequence}
For any $\hsic_0$, there exists a \KS $\{\fm_{l^{\circ}}\}_{l=1}^L$ parameterized by a set of weights $W_l$ and a set of bandwidths $\sigma_l$ such that:
\begin{enumerate}[label=\Roman*.]
    \item 
    as $L \rightarrow \infty$, the \RS converges to the global optimum, that is 
    \begin{equation}
        \lim_{L \rightarrow \infty}  \hsic_L = \hsic^*,
    \end{equation}   
    \item 
    the convergence is strictly monotonic where 
    \begin{equation}
    \hsic_{l} > \hsic_{l-1} \quad \forall l \ge 1.
    \end{equation}   
\end{enumerate}
\end{theorem}
\end{adjustwidth}

\textbf{On Generalization. } The ISM algorithm provides some insight into generalization. While the HSIC objective employs an infinitely wide network, \citet{ma2019hsic} have experimentally observed that HSIC can generalize even without any regularizer. We ask theoretically, what makes the HSIC objective special?  Recently, \citet{poggio2020complexity} have proposed that  traditional MLPs generalize  because gradient methods implicitly regularize the normalized weights. We discovered a similar impact ISM has on HSIC, i.e., the objective can be reformulated to isolate out $n$ functions $[D_1(W_l), ..., D_n(W_l)]$ that act as a penalty term during optimization. Let $\cS_i$ be the set of samples that belongs to the $i_{th}$ class and let $\cS^c_i$ be its complement, then each function $D_i(W_l)$ is defined as
\begin{equation}
    \label{eq:penalty_term}
    D_i(W_l) = 
    \frac{1}{\sigma^2}
    \sum_{j \in \cS_i}
    \Gij \kf_{W_l}(r_i,r_j)
    -
    \frac{1}{\sigma^2}
    \sum_{j \in \cS^c_i}
    |\Gij| \kf_{W_l}(r_i,r_j).
\end{equation}

Note that $D_i(W_l)$ is simply \eq{eq:similarity_hsic} for a single sample scaled by $\frac{1}{\sigma^2}$. Therefore, as we identify better solutions for $W_l$, this leads to an increase and decrease of $\kf_{W_l}(r_i,r_j)$ associated with $\cS_i$ and $\cS^c_i$ in \eq{eq:penalty_term}, thereby increasing the size of the penalty term  $D_i(W_l)$.  To appreciate how $D_i(W_l)$ penalizes $\hsic$, we propose an equivalent formulation with its derivation in App.~B. 
\begin{adjustwidth}{0.2cm}{0.0cm}
\vspace{3px}
\begin{theorem}
\label{thm:regularizer}
\eq{eq:main_obj} is equivalent to 
\begin{equation}
    \max_{W_l}
    \sum_{i,j}
    \frac{\Gij}{\sigma^2}
    \ISMexp
    (r_i^TW_lW_l^Tr_j)
    -
    \sum_{i}
    D_i(W_l)
    ||W_l^Tr_i||_2.
    \label{eq:generalization_formulation}
\end{equation}
\end{theorem}
\end{adjustwidth}
Based on Thm.~\ref{thm:regularizer}, $D_i(W_l)$ adds a negative cost to the sample norm in IDS, $||W_l^Tr_i||_2$, describing how ISM implicitly regularizes HSIC. As a better $W_l$ attempts to improve the objective, it simultaneously imposes a heavier penalty on \eq{eq:generalization_formulation} where the overall $\hsic$ may actually decrease.

%
%
%


\section{Experiments}
\textbf{Datasets. }
We confirm the theoretical properties of \kc using three synthetic (Random, Adversarial and Spiral) and five popular UCI benchmark datasets: wine, cancer, car, divorce, and face ~\cite{Dua:2017}. To test the flexibility of \kc, we design the Adversarial dataset to be highly complex, i.e., the samples pairs in $\cS^c$ are significantly closer than sample pairs in $\cS$. We next designed a Random dataset with completely random labels. All datasets are included along with the source code in the supplementary, and their comprehensive download link and statistics are in App.D. 

\textbf{Evaluation Metrics and Settings. } To evaluate the central claim that MLPs can be solved greedily, we report $\hsic^*$ at convergence along with the training/test accuracy for each dataset. Here, $\hsic^*$ is normalized to the range between 0 to 1 using the method proposed by \citet{cortes2012algorithms}. Since \kc at convergence is itself a feature map, we evaluate the network output quality with the Cosine Similarity Ratio ($\csr$). The $\langle \fm(x_i), \fm(x_j) \rangle$ for $i,j$ pairs in $\cS$ and $\cS^c$ should be 1 and 0, yielding $\csr = 0$. The equations for $\hsic^*$ and $\csr$ are

    \begin{equation}
        \hsic^* = \frac{\hsic(\fm(X),Y)}{\sqrt{\hsic(\fm(X),\fm(X)) \hsic(Y,Y)}} 
        \quad \text{and} \quad
        \csr = \frac
        {\sum_{i,j \in \mathcal{S}^c} \langle \fm(x_i), \fm(x_j) \rangle}
         {\sum_{i,j \in \mathcal{S}} \langle \fm(x_i), \fm(x_j) \rangle}.
    \end{equation}    
      The RFF length is set to 300 for all datasets and the $\sigma_l$ that maximizes $\hsic^*$ is chosen. The convergence threshold for \RS is set at $\hsic_l > 0.99$. The network structures discovered by ISM for every dataset are recorded and provided in App.~E.
      The MLPs that use $\mse$ and $\ce$ have weights initialized via the Kaiming method \cite{he2015delving}. All datasets are centered to 0 and scaled to a standard deviation of 1. All sources are written in Python using Numpy, Sklearn and Pytorch \cite{numpy,sklearn_api,paszke2017automatic}. All experiments were conducted on an Intel Xeon(R) CPU E5-2630 v3 @ 2.40GHz x 16 with 16 total cores. 

\textbf{Experimental Results. }
We conduct 10-fold cross-validation across all 8 datasets and reported their mean and the standard deviation for all key metrics. The random and non-random datasets are visually separated. Once our model is trained and has learned its structure, we use the same depth and width to train 2 additional MLPs via SGD, where instead of HSIC, $\mse$ and $\ce$ are used as the empirical risk. The results are listed in Table~\ref{table:main} with the best outcome in bold.

Can \RS be optimized greedily? The $\hsic^*$ column in Table~\ref{table:main} consistently reports results that converge near its theoretical maximum value of 1, thereby corroborating with Thm.~\ref{thm:hsequence}. As we discover a better kernel, \kc discovers a mapping that separates the dataset into distinguishable clusters, producing high training accuracies as $\hsic_l \rightarrow \hsic^*$. 
Will our network generalize? 
Since smooth mappings are associated with better generalization, we also report the smallest $\sigma$ value used for each network to highlight the smoothness of $\fm$ learned by ISM. Correspondingly, with the exception of the two random datasets, our test accuracy consistently performed well across all datasets. While we cannot definitively attribute the impressive test results to Thm.~\ref{thm:regularizer}, the experimental evidence appears to be aligned with its implication. 

\begin{table}[h]
\tiny
\centering
\setlength{\tabcolsep}{5.0pt}
\renewcommand{\arraystretch}{1.2}
\begin{tabular}{|c|c|c|c|c|c|c|c|c|c|c|c|}
	\hline
	 & obj &
		 $\sigma$  $\uparrow$&
		 $L$ $\downarrow$&
		 Train Acc $\uparrow$&
		 Test Acc $\uparrow$&
		 Time(s) $\downarrow$&
		 $\hsic^*$ $\uparrow$&
		 $C$ $\downarrow$\\
		\hline 
	\parbox[t]{2mm}{\multirow{3}{*}{\rotatebox[origin=c]{90}{random}}} &
	 $\hsic$ &
		 0.38 &
		 \black{3.30 $\pm$ 0.64} &
		 \textbf{\black{1.00 $\pm$ 0.00}} &
		 \black{0.38 $\pm$ 0.21} &
		 \textbf{\black{0.40 $\pm$ 0.37}} &
		 \textbf{\black{1.00 $\pm$ 0.01}} &
		 \textbf{\black{0.00 $\pm$ 0.06}} \\
	 & $\ce$ &
		 - &
		 \black{3.30 $\pm$ 0.64} &
		 \textbf{\black{1.00 $\pm$ 0.00}} &
		 \black{0.48 $\pm$ 0.17} &
		 \black{25.07 $\pm$ 5.55} &
		 \textbf{\black{1.00 $\pm$ 0.00}} &
		 \textbf{\black{0.00 $\pm$ 0.00}} \\
	 & $\mse$ &
		 - &
		 \black{3.30 $\pm$ 0.64} &
		 \black{0.98 $\pm$ 0.04} &
		 \textbf{\black{0.63 $\pm$ 0.21}} &
		 \black{23.58 $\pm$ 8.38} &
		 \black{0.93 $\pm$ 0.12} &
		 \black{0.04 $\pm$ 0.04} \\
		\hline 
	\parbox[t]{2mm}{\multirow{3}{*}{\rotatebox[origin=c]{90}{adver}}} &
    $\hsic$ &
		 0.5 &
		 \black{3.60 $\pm$ 0.92} &
		 \textbf{\black{1.00 $\pm$ 0.00}} &
		 \textbf{\black{0.38 $\pm$ 0.10}} &
		 \textbf{\black{0.52 $\pm$ 0.51}} &
		 \textbf{\black{1.00 $\pm$ 0.00}} &
		 \textbf{\black{0.01 $\pm$ 0.08}} \\
	 & $\ce$ &
		 - &
		 \black{3.60 $\pm$ 0.92} &
		 \black{0.59 $\pm$ 0.04} &
		 \black{0.29 $\pm$ 0.15} &
		 \black{69.54 $\pm$ 24.14} &
		 \black{0.10 $\pm$ 0.07} &
		 \black{0.98 $\pm$ 0.03} \\
	 & $\mse$ &
		 - &
		 \black{3.60 $\pm$ 0.92} &
		 \black{0.56 $\pm$ 0.02} &
		 \black{0.32 $\pm$ 0.20} &
		 \black{113.75 $\pm$ 21.71} &
		 \black{0.02 $\pm$ 0.01} &
		 \black{0.99 $\pm$ 0.02} \\
		\hline 
		\hline 
	\parbox[t]{2mm}{\multirow{3}{*}{\rotatebox[origin=c]{90}{spiral}}} &
	 $\hsic$ &
		 0.46 &
		 \black{5.10 $\pm$ 0.30} &
		 \textbf{\black{1.00 $\pm$ 0.00}} &
		 \textbf{\black{1.00 $\pm$ 0.00}} &
		 \textbf{\black{0.87 $\pm$ 0.08}} &
		 \black{0.98 $\pm$ 0.01} &
		 \black{0.04 $\pm$ 0.03} \\
	 & $\ce$ &
		 - &
		 \black{5.10 $\pm$ 0.30} &
		 \textbf{\black{1.00 $\pm$ 0.00}} &
		 \textbf{\black{1.00 $\pm$ 0.00}} &
		 \black{11.59 $\pm$ 5.52} &
		 \textbf{\black{1.00 $\pm$ 0.00}} &
		 \textbf{\black{0.00 $\pm$ 0.00}} \\
	 & $\mse$ &
		 - &
		 \black{5.10 $\pm$ 0.30} &
		 \textbf{\black{1.00 $\pm$ 0.00}} &
		 \black{0.99 $\pm$ 0.01} &
		 \black{456.77 $\pm$ 78.83} &
		 \textbf{\black{1.00 $\pm$ 0.00}} &
		 \black{0.40 $\pm$ 0.01} \\
		\hline 
	\parbox[t]{2mm}{\multirow{3}{*}{\rotatebox[origin=c]{90}{wine}}} &
	 $\hsic$ &
		 0.47 &
		 \black{6.10 $\pm$ 0.54} &
		 \black{0.99 $\pm$ 0.00} &
		 \textbf{\black{0.97 $\pm$ 0.05}} &
		 \textbf{\black{0.28 $\pm$ 0.04}} &
		 \black{0.98 $\pm$ 0.01} &
		 \black{0.04 $\pm$ 0.03} \\
	 & $\ce$ &
		 - &
		 \black{6.10 $\pm$ 0.54} &
		 \textbf{\black{1.00 $\pm$ 0.00}} &
		 \black{0.94 $\pm$ 0.06} &
		 \black{3.30 $\pm$ 1.24} &
		 \textbf{\black{1.00 $\pm$ 0.00}} &
		 \textbf{\black{0.00 $\pm$ 0.00}} \\
	 & $\mse$ &
		 - &
		 \black{6.10 $\pm$ 0.54} &
		 \textbf{\black{1.00 $\pm$ 0.00}} &
		 \black{0.89 $\pm$ 0.17} &
		 \black{77.45 $\pm$ 45.40} &
		 \textbf{\black{1.00 $\pm$ 0.00}} &
		 \black{0.49 $\pm$ 0.02} \\
		\hline 
	\parbox[t]{2mm}{\multirow{3}{*}{\rotatebox[origin=c]{90}{cancer}}} &
	 $\hsic$ &
		 0.39 &
		 \black{8.10 $\pm$ 0.83} &
		 \black{0.99 $\pm$ 0.00} &
		 \textbf{\black{0.97 $\pm$ 0.02}} &
		 \textbf{\black{2.58 $\pm$ 1.07}} &
		 \black{0.96 $\pm$ 0.01} &
		 \black{0.02 $\pm$ 0.04} \\
	 & $\ce$ &
		 - &
		 \black{8.10 $\pm$ 0.83} &
		 \textbf{\black{1.00 $\pm$ 0.00}} &
		 \textbf{\black{0.97 $\pm$ 0.01}} &
		 \black{82.03 $\pm$ 35.15} &
		 \textbf{\black{1.00 $\pm$ 0.00}} &
		 \textbf{\black{0.00 $\pm$ 0.00}} \\
	 & $\mse$ &
		 - &
		 \black{8.10 $\pm$ 0.83} &
		 \textbf{\black{1.00 $\pm$ 0.00}} &
		 \textbf{\black{0.97 $\pm$ 0.03}} &
		 \black{151.81 $\pm$ 27.27} &
		 \textbf{\black{1.00 $\pm$ 0.00}} &
		 \textbf{\black{0.00 $\pm$ 0.0}} \\
		\hline 
	\parbox[t]{2mm}{\multirow{3}{*}{\rotatebox[origin=c]{90}{car}}} &
	 $\hsic$ &
		 0.23 &
		 \black{4.90 $\pm$ 0.30} &
		 \textbf{\black{1.00 $\pm$ 0.00}} &
		 \textbf{\black{1.00 $\pm$ 0.01}} &
		 \textbf{\black{1.51 $\pm$ 0.35}} &
		 \black{0.99 $\pm$ 0.00} &
		 \black{0.04 $\pm$ 0.03} \\
	 & $\ce$ &
		 - &
		 \black{4.90 $\pm$ 0.30} &
		 \textbf{\black{1.00 $\pm$ 0.00}} &
		 \textbf{\black{1.00 $\pm$ 0.00}} &
		 \black{25.79 $\pm$ 18.86} &
		 \textbf{\black{1.00 $\pm$ 0.00}} &
		 \textbf{\black{0.00 $\pm$ 0.00}} \\
	 & $\mse$ &
		 - &
		 \black{4.90 $\pm$ 0.30} &
		 \textbf{\black{1.00 $\pm$ 0.00}} &
		 \textbf{\black{1.00 $\pm$ 0.00}} &
		 \black{503.96 $\pm$ 116.64} &
		 \textbf{\black{1.00 $\pm$ 0.00}} &
		 \black{0.40 $\pm$ 0.00} \\
		\hline 
	\parbox[t]{2mm}{\multirow{3}{*}{\rotatebox[origin=c]{90}{face}}} &
	 $\hsic$ &
		 0.44 &
		 \black{4.00 $\pm$ 0.00} &
		 \textbf{\black{1.00 $\pm$ 0.00}} &
		 \textbf{\black{0.99 $\pm$ 0.01}} &
		 \textbf{\black{0.78 $\pm$ 0.08}} &
		 \black{0.97 $\pm$ 0.00} &
		 \black{0.01 $\pm$ 0.00} \\
	 & $\ce$ &
		 - &
		 \black{4.00 $\pm$ 0.00} &
		 \textbf{\black{1.00 $\pm$ 0.00}} &
		 \black{0.79 $\pm$ 0.31} &
		 \black{23.70 $\pm$ 8.85} &
		 \textbf{\black{1.00 $\pm$ 0.00}} &
		 \textbf{\black{0.00 $\pm$ 0.00}} \\
	 & $\mse$ &
		 - &
		 \black{4.00 $\pm$ 0.00} &
		 \black{0.92 $\pm$ 0.10} &
		 \black{0.52 $\pm$ 0.26} &
		 \black{745.17 $\pm$ 281.56} &
		 \black{0.94 $\pm$ 0.07} &
		 \black{0.72 $\pm$ 0.01} \\
		\hline 
	\parbox[t]{2mm}{\multirow{3}{*}{\rotatebox[origin=c]{90}{divorce}}} &
	 $\hsic$ &
		 0.41 &
		 \black{4.10 $\pm$ 0.54} &
		 \black{0.99 $\pm$ 0.01} &
		 \black{0.98 $\pm$ 0.02} &
		 \textbf{\black{0.71 $\pm$ 0.41}} &
		 \black{0.99 $\pm$ 0.01} &
		 \textbf{\black{0.00 $\pm$ 0.05}} \\
	 & $\ce$ &
		 - &
		 \black{4.10 $\pm$ 0.54} &
		 \textbf{\black{1.00 $\pm$ 0.00}} &
		 \textbf{\black{0.99 $\pm$ 0.02}} &
		 \black{2.62 $\pm$ 1.21} &
		 \textbf{\black{1.00 $\pm$ 0.00}} &
		 \textbf{\black{0.00 $\pm$ 0.00}} \\
	 & $\mse$ &
		 - &
		 \black{4.10 $\pm$ 0.54} &
		 \textbf{\black{1.00 $\pm$ 0.00}} &
		 \black{0.97 $\pm$ 0.03} &
		 \black{47.89 $\pm$ 24.31} &
		 \textbf{\black{1.00 $\pm$ 0.00}} &
		 \textbf{\black{0.00 $\pm$ 0.01}} \\
	\hline
\end{tabular}
\vspace{3pt}
\caption{Each dataset contains 3 rows comparing the greedily trained \kc using $\hsic$ against traditional MLPs of the same size trained using MSE and CE via SGD given the same network width and depth. The best results are in bold with $\uparrow/\downarrow$ indicating larger/smaller values preferred.}
\label{table:main}
\end{table}

Since Thm.~\ref{thm:hsequence} also claims that we can achieve $\hsic^* - \hsic_l < \delta$ in finite number of layers, we include in Table~\ref{table:main} the average length of the \RS ($L$). The table suggests that the \RS converges quickly with 9 layers as the deepest network. 
The execution time for each objective is also recorded for reference in Table~\ref{table:main}. Since \kc can be solved via a single forward pass while SGD requires many iterations of BP, \kc should be faster. The Time column of Table~\ref{table:main} confirmed this expectation by a wide margin. The biggest difference can be observed by comparing the face dataset, $\hsic$ finished with 0.78 seconds while $\mse$ required 745 seconds; that is almost 1000 times difference. While the execution times reflect our expectation, techniques that vastly accelerate kernel computations \cite{wang2019exact,rudi2017falkon} would be required for larger datasets. Lastly, \kc induces low $\csr$ as shown in Table~\ref{table:main}, implying that samples in $\cS$ and $\cS^c$ are being pulled together and pushed apart in RKHS via the angular distance.

\clearpage
\bibliography{reference}
\bibliographystyle{unsrtnat}

\clearpage
\newpage
\onecolumn

\end{document}